# HIERARCHICAL PIXEL CLUSTERING FOR IMAGE SEGMENTATION

M. Kharinov

St. Petersburg Institute for Informatics and Automation of RAS,
St. Petersburg, Russia

*e*-mail: khar@iias.spb.su

*In the paper a piecewise constant image approximations of sequential number of pixel clusters or segments are treated. A majorizing of optimal approximation sequence by hierarchical sequence of image approximations is studied. Transition from pixel clustering to image segmentation by reducing of segment numbers in clusters is provided. Algorithms are proved by elementary formulas.*

**Introduction**

The paper focuses on the domain of image segmentation by optimal approximations that minimally differ from the image of $N$ pixels in the standard deviation $\sigma$ or total squared error $E = N\sigma^2$. Although related approaches, namely Otsu methods [1, 2], *K*-means method [3] and Mumford-Shah model [4–7] have a long history, the opportunities of minimizing of the total squared error $E$ are far from being exhausted, especially, in the task of multiple optimization for each number of pixel clusters or, in particular, connected image segments. In this task Otsu's multi-thresholding [2] provides an accurate but incomplete solution for clustering of pixels. Mumford-Shah model [4–7] provides a complete sequence of image partitions into each number of segments, but minimizing effect is poor. *K*-means method for image segmentation is too heuristic to provide any of mentioned two requirements, but it can be advanced for application in conjunction with Otsu method and Mumford-Shah model [8].

To solve the task of multiple optimization without any difficulties we use a special data structure of Sleator-Tarjan dynamic trees [9] that essentially optimizes the computing, but does not affect the obvious meaning of algorithms. Therefore, to avoid the cumbersome details of implementation, here we address rather to motivation of solutions and do not dwell on the software that supports the fast generation, storing in the available RAM and effective transformations of pixel clusters in a computer memory.

To substantiate the study of segmentation results without appealing to the subsequent detection of a priori specified objects, we have calculated the optimal and nearly optimal approximations for the simplest examples of real images [10]. These proved important for the formulation of the problem caused by two challenges.

**1. Problem statement**

The first challenge is that a sequence of optimal image approximations in general case is not hierarchical [10]. But just hierarchical sequence of approximations is quite accessible for computational optimization. Therefore, there arises the problem of majorizing of none-hierarchical optimal approximation sequence by quasioptimal hierarchical sequence of approximations, which don't significantly differ from the optimal ones in total squared error $E$ or standard deviation $\sigma$.

The second challenge is illustrated by two-level approximations in Fig. 1.

Leftmost in Fig.1 is the original image, central is the optimal approximation with two intensities obtained by conventional Otsu method, providing the minimum of standard deviation of approximation from the image. Leftmost is the nearly optimal approximation of original image by two segments. The values of standard deviation are written under the approximations.

Comparing the two approximations in Fig. 1, it is easy to notice that in nearly optimal approximation the segments of optimal approximation are connected to each other by natural or artificial coupling elements one pixel wide. The calculations of such coupling elements are unstable. With increasing intensity resolution the contribution of coupling elements in the total squared error tends to zero. Then, none-literal «virtual» coupling elements turn out to be preferable, and we come to utilization of pixel clusters instead of less general connected segments.

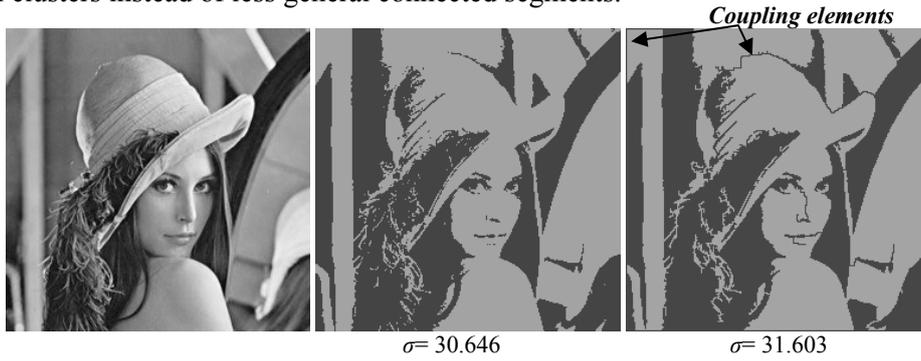

$\sigma= 30.646$  $\sigma= 31.603$

**Fig. 1.** Two-level optimal and nearly optimal approximations of the standard Lenna image.

In this case, the problem is to obtain the complete sequence of the quasioptimal image approximations by pixel clusters with a limited number of segments. The decisive consideration to treat clusters of connected and disconnected pixels instead of less common connected segments consists in that for clusters, besides a sole merging operation, two additional operations, which don't cause withdrawal from their sets, are introduced.

## 2. Elementary formulas for total squared error minimizing

To optimize image approximations by the total squared error $E$ or standard deviation $\sigma$, we use three operations with pixel clusters, namely, merging, splitting and correction, described by the following formulas.

Let $I_1$ and $I_2$ be the average intensities for clusters 1 and 2, respectively. Let $n_1$ be the number of pixels in the cluster 1 and $n_2$ be the number of pixels in the cluster 2. Then the increment $\Delta E_{merge}$ of the total squared error $E$ caused by the merging of specified clusters along with reduction of the number of clusters per unit is given by the formula:

$$\Delta E_{merge} = \frac{(I_1 - I_2)^2}{\frac{1}{n_1} + \frac{1}{n_2}} \geq 0 \qquad (1)$$

Just the increment $\Delta E_{merge}$ is minimized in the version [4] of Mumford-Shah model [5], however, in application to connected segments. In the version [6], the appropriate formula differs by an additive term, and in FLSA version [7] by a multiplicative factor to take into account the total length of the boundaries between the segments (clusters of connected pixels).

Let's write down the formula for splitting of the cluster 1, when its $k < n_1$ pixels with average intensity $I$ initiate a new cluster. In this case, the cluster 1 is split into two clusters of $k$ and complementary $n_1 - k$ pixels, and cluster splitting is accompanied with increase of the cluster number per unit along with a non-positive increment $\Delta E_{split}$ of the total squared error:

$$\Delta E_{split} = -\frac{(I - I_1)^2}{\frac{1}{k} - \frac{1}{n_1}} \leq 0 \qquad (2)$$

For cluster splitting it is important that in (2) a predetermined set of pixel clusters is expected, which obtained, say, by cluster merging using (1). So, the extended set of clusters is taken into account when splitting. Another feature of algorithms based on the formula (2), is the update of the hierarchy of clusters, which is performed for each of the nested clusters, treated both as the individual images.

The composition of splitting and merging of clusters induces a correction operation without changing the number of clusters, which is accompanied with an increment $\Delta E_{correct}$ of the total squared error:

$$\Delta E_{correct} = \frac{\|I - I_2\|^2}{\frac{1}{k} + \frac{1}{n_2}} - \frac{\|I - I_1\|^2}{\frac{1}{k} - \frac{1}{n_1}}, \tag{3}$$

where the negative term in (3) describes the increment of the total squared error $E$, caused by converting of $k$ pixels from cluster 1 into a separate cluster, and the first term in (3) describes the increment of $E$ caused by merging of the initiated cluster with the cluster 2, in accordance with (1) and (2).

A notable feature of (3) is that by its simplifying the method *K*-means is derived [8]. Applying (3) precisely, we have proposed for the clustering of pixel sets a more accurate method [10], which provides a calculation of a complete sequence of optimal image approximations that are treated in multi-threshold Otsu method [2]. Another feature of formula interpretation in this paper is the application of (3) for generation of a hierarchy of pixel clusters, unlike overlapping partitioning, as in [8, 10].

### 3. Quasioptimal image approximations

To produce the sequences of quasioptimal approximations, we follow the principle of dichotomous division of the cluster into two subclusters independently of the others, treating each cluster as a separate image. In this way the splitting of non-uniform clusters containing the different pixels is performed, while clusters of all identical pixels are treated as indivisible or elementary. To avoid analysis of cluster repetitions the computation of hierarchical sequence of quasioptimal image approximations, containing 1, 2, 3, ... clusters of pixels, is performed in two stages. At the first stage so called «compact» invariant representation, which specifies the sequence of partitions of the image pixels into 1, 2, 4, 8 ... clusters, is calculated. At the second stage, a compact representation is expanded into a sequence of approximations with successively increasing numbers of clusters, so as to provide the maximal decrease of the total squared error $E$ or standard deviation $\sigma$.

In an obvious way the sequence of quasioptimal image approximations is obtained by splitting of the non-uniform clusters according to conventional histogram Otsu method [2], wherein the threshold intensity value is found using exhaustive search, from the condition of maximum decrease of the total squared error $E$.

In a more complicated algorithm, the quasioptimal approximations are generated according to the formula (2), using a hierarchy of clusters that previously generated in bottom-up strategy by brute-force implementation of formula (1) for the minimizing of total squared error $E$ over all pairs of clusters. As we have established experimentally, the results of calculations according (1) and (2) coincide with each other. Moreover, the equivalent hierarchy of none-uniform pixel clusters can be generated in the simplest algorithm [11], if we exchange the heuristic criterion [11] of merging of successive histogram bins by criterion $\Delta E_{merge} = \min$, where $\Delta E_{merge}$ is detailed in (1).

Characteristic feature of optimal and quasioptimal image approximations with $g = 1, 2, 3, 4...$ number of pixel clusters is that the corresponding sequence $E_1 \geq E_2 \geq, ..., \geq E_g$ of values of the total squared error $E$ is convex:

$$E_i \leq \frac{E_{i-1} + E_{i+1}}{2}, \quad i = 2, 3, ..., g-1. \tag{4}$$

Convexity property (4) holds also for a compact sequence of quasioptimal image approximations with $g = 1, 2, 4, 8...$ clusters. Thus, the quasioptimal approximations preserve a convexity property of the optimal image approximations.

Fig. 2 shows the image approximations with two, three and four pixel clusters, visualized by the same number of average intensities. Comparing Fig. 2 with Fig. 1, it is easy to notice the visual difference of approximations that is less expressed in the values of $E$ or $\sigma$.

### 4. Correcting image approximations by segment number

The main limitation of quasioptimal approximations Fig. 2 in image segmentation task is a sharp increase in the number of connected segments with increasing number of pixel clusters. Howev-er, this limitation is overcome due to the reducing of the number of segments in the image approximation that is performed as an additional step in algorithm of generation of quasioptimal approximations by cluster splitting into two subclusters.

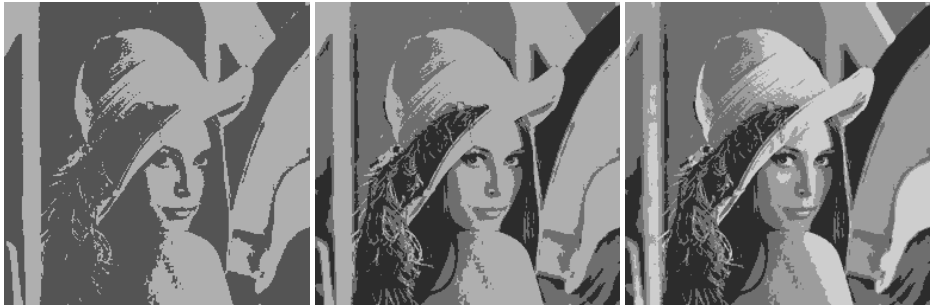

**Fig. 2.** Quasioptimal image approximations with 2, 3, 4 pixel clusters.

Reducing the number of segments is performed by reclassifying pixels of non-isolated segment from donor subcluster to the acceptor subcluster that are selected to minimize the increment of the total squared error $\Delta E_{correct} = \min$, where $\Delta E_{correct}$ is detailed in (3).

Fig. 3 illustrates the quality $\sigma$ of approximations of the standard image, depending on the number of clusters $g$ shown in the range from one to one thousand.

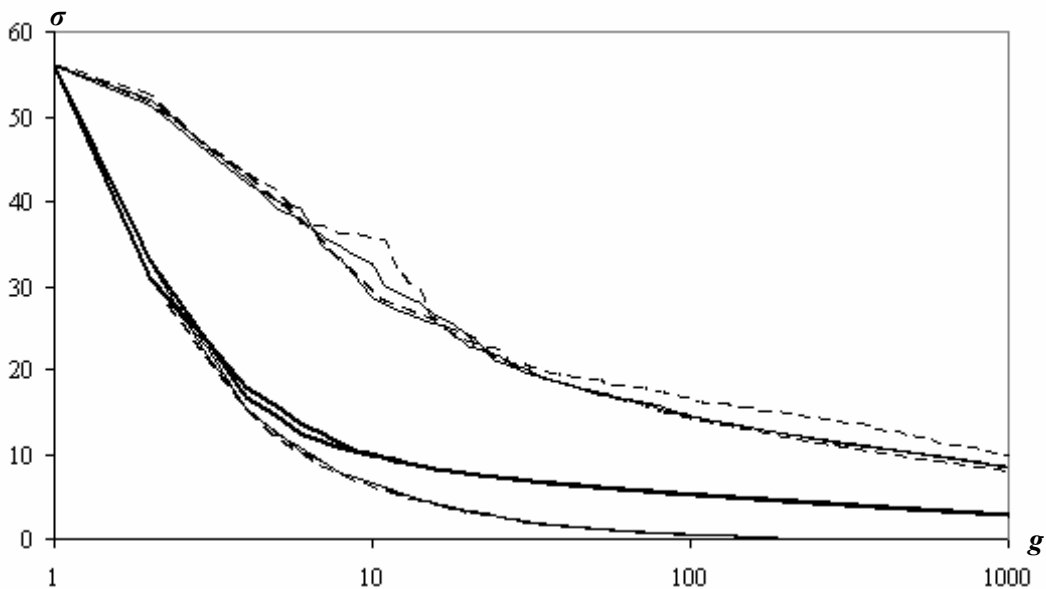

**Fig. 3.** Standard deviation $\sigma$ depending on the number of clusters $g$ (logarithmic scale).

In Fig. 3 the pairs of intertwining solid curves describe the sequence of approximations obtained by top-down clustering according Otsu method and bottom-up clustering by iterative merging. Lower boundary dotted curve marks the optimal approximations with successive number of pixel clusters. The pair of almost merged solid curves just above this dashed curve describes two majorizing sequences of quasioptimal image approximations. Upper dashed curve describes image approximations with successive number of connected segments according to FLSA version [7] of Mumford-Shah model, and just below dashed curve corresponds to the version [4] of Mumford-Shah model [5].

The uppermost pair of intertwined solid curves describes the sequence of approximations obtained in discussed bottom-up and top-down algorithms, along with the reduction of the number of segments in each subcluster to one. In this case the sequence of image approximation with sequential number of segments is generated, as in the Mumford-Shah model. Intermediate pair of bold curves describes a sequence of approximations obtained by reduction of the number of segments, which terminates under certain stopping condition. In discussed particular case as a stopping condition was taken a condition of uniqueness of average intensity of each segment that causes three-five times reduction in the segment number, compared to quasioptimal approximations. For a variety of stopping conditions the area between the curves in Fig. 3 becomes available that extends the capabilities of image segmentation via pixel clustering.

**Conclusion**

Thus, the quasioptimal image approximations Fig.1 may be reproduced by the histogram algorithm [11], wherein the «distance between the clusters», i.e. the product of intra-class and inter-class variances, is to be replaced by $\Delta E_{merge}$ from (1)[1]. As we have established experimentally, this algorithm provides the minimization, which remains valid for the general case of merging of any cluster pairs. Along the way, we have created software for the joint analysis of global and local pixel features, which came in handy to develop a clustering method by reduction of the number of segments.

It should be noted that in addition to the availability of computing, the quasioptimal image approximations have one more remarkable advantage, compared to optimal approximations. Concretely, the quasioptimal approximations are easily converted into invariant image representations [12] that don't depend on the linear transformations of pixel intensities. This topic should be discussed in the following papers.

---

[1] The optimal approximations of the standard Lenna image, which may be desired for comparison, are available at http://oogis.ru/content/view/107/42/lang,ru/.